\documentclass[runningheads,a4paper]{llncs}

\usepackage{amssymb}
\usepackage[utf8]{inputenc}
\usepackage{graphicx}
\usepackage{listings}
\usepackage{parcolumns}
\usepackage[english]{babel}
\usepackage{blindtext}

\usepackage{url}
\urldef{\maila}\path|{hos,victordantas2}@edu.unifor.br, vasco@unifor.br,|
\urldef{\mailb}\path|pinhep@rpi.edu, dlm@cs.rpi.edu|

\begin{document}

\mainmatter  

\title{From Data to City Indicators:\\A Knowledge Graph for Supporting\\Automatic Generation of Dashboards}

\titlerunning{A Knowledge Graph for Supporting Automatic Generation of Dashboards}

%
%
\author{Henrique Santos\inst{1}%
\and Victor Dantas\inst{1}%
\and Vasco Furtado\inst{1}%
\and\\Paulo Pinheiro\inst{2}%
\and Deborah L. McGuinness\inst{2}}
\authorrunning{Henrique Santos et al.}

\institute{Universidade de Fortaleza, Fortaleza, CE, Brazil
\and
Rensselaer Polytechnic Institute, Troy, NY, U.S.A.
\\
\maila
\\
\mailb}

%
%

\toctitle{From Data to City Indicators: A Knowledge Graph for Supporting Automatic Generation of Dashboards}
\tocauthor{Henrique Santos et al.}
\maketitle

\begin{abstract}
In the context of Smart Cities, indicator definitions have been used to calculate values that enable the comparison among different cities. The calculation of an indicator values has challenges as the calculation may need to combine some aspects of quality while addressing different levels of abstraction. Knowledge graphs (KGs) have been used successfully to support flexible representation, which can support improved understanding and data analysis in similar settings. This paper presents an operational description for a city KG, an indicator ontology that support indicator discovery and data visualization and an application capable of performing metadata analysis to automatically build and display dashboards according to discovered indicators. We describe our implementation in an urban mobility setting.
\end{abstract}

\section{Introduction}

While a single agreed upon definition of a smart city may be elusive, many definitions, if not most definitions include some technology and infrastructure that provide a high quality of life for its residents. Determining a desirable quality of life often includes evaluation of city qualities such as: sustainability, safety, inclusiveness, walkability, creativity, and innovation. Cities with high scores on these qualities are often judged as being desirable places to live. Achieving the capability of assessing any of these or other desirable qualities, however, requires two key components: accessing and understanding city's data. Consequently, a city's ability to produce and share relevant data that can be understood and used by a broad range of diverse stakeholders is critical for evaluating and comparing cities and can be viewed as key indicator of a Smart City as well as the ability to derive knowledge from city’s data and further use it to power innovation.

Governments are increasingly sharing city data, often with the goal of promoting innovation via societal participation with the use of data. In the context of data sharing, different categories of stakeholders may be identified: designers and software developers may use data to produce public services through the use of web and mobile applications; scientists may produce elaborate analysis and studies about the cities; public officers may use the data to improve city administration using data-based decision-making techniques; journalists may use open data to produce more reliable, factually-based and attractive news. The use of knowledge graphs (KGs) as a way of better understanding and analyzing data has proven successful in many cases\cite{biega_inside_2013}\cite{dong_knowledge_2014}\cite{hoffart_yago2:_2011}\cite{singhal_introducing_????}. They are not simply linked data using an RDF model; they also provide support for knowledge management including explicit provenance encoding capabilities, entity description encodings, potential to connect to and leverage reasoners and so forth.

To obtain measured values for characterizing city's properties, some approaches \cite{fox_polisgnosis_2015}\cite{fox_role_2015}\cite{iso_sustainable_2014} have made use of the development and calculation of city indicators. Indicators are metrics that one can use to assess the city level of maturity in a certain field of interest. More than that, well-defined indicators enable the comparison among different cities so one can determine when one city appears to be doing better than another city with respect to certain criteria. However, robust, reusable, and precise calculation plans for indicators have challenges. For example compound indicators require combinations of data that may be unavailable and those data need to be modeled in enough detail so that indicator calculating systems (and humans) can understand enough to know when data is comparable and may be combined. Further enough information about provenance needs to be available so that trust can be ascertained.

This paper tackles the challenge of calculating indicator values from (raw) data, describing work with both city indicators and KGs for city data as a way to automatically build and display dashboards that can be used by a wide range of users in city comparisons. The proposed KG uses OWL ontologies that describe concepts and relations regarding sensing infrastructure, provenance, data acquisition activities, indicators and city entities themselves. Once built, the KG (or a subset of it) can be serialized in the Contextualized CSV format (CCSV\cite{santos_contextual_2015} - a format that conveys both data and associated metadata) while a reasoner performs inferences to discover indicators inside the indicator ontologies that are suited for the serialized data. Discovered indicators are then serialized themselves in Turtle format. Both serializations are presented to a dashboard generating application, which performs metadata analysis to automatically build and display dashboards according to the discovered indicators. The three main contributions of this paper are (i) the city KG description that enables transparent and explainable indicator values; (ii) the Indicator ontology that can support dashboard visualization; and (iii) a dashboard generating application that works with knowledge from KGs. The rest of this paper is organized as it follows. The next section introduces current approaches to KGs, city indicators, city modeling and data annotation. In Section 3, the proposed KG is defined alongside its ontologies, modeling decisions and serialization process. Section 4 describes the dashboard generating application and its metadata analysis that supports building and displaying dashboards in the context of urban mobility, Section 5 concludes and discusses future plans.

\section{Related Work}

Recently, the Knowledge Graph (KG) phrase has been used to define large collections of structured data in a meaningful way. Being more than simply linked data, the semantics encoded in a KG enables tasks that may be challenging in simple linked data RDF models. Metadata faceting, provenance tracking and context-awareness are examples of enhanced features that KGs can support. The term gained popularity with Google KG\cite{singhal_introducing_????} in a effort to merge Freebase\cite{bollacker_freebase:_2008} (which also may be considered  a KG), Wikipedia and the CIA World Factbook\footnote{\url{https://www.cia.gov/library/publications/the-world-factbook}} augmented with their search engine’s queries and results. Academic KGs are also available including YAGO\cite{biega_inside_2013}\cite{hoffart_yago2:_2011} and DBpedia\cite{auer_dbpedia:_2007}.

\subsection{City Modeling}

The process of modeling a city is complex. The intrinsic complexity of interactions between city entities make it very difficult to map relevant sets of dynamic aspects that are often used to characterize a city. Moreover, these entity interactions, along with the numerous entities and processes, differ from one city to another. Thus, the process of modeling the city is typically use-case centered, where the modeling is performed towards a specified goal. This approach, hence, streamlines the process, identifying which characteristics need to be modeled. The work in \cite{zhao_toward_2015} proposes a core conceptual model for the Domain Knowledge Model of a Smart City, which originally involves multiple domains and cities. The proposed work aims to support cross-domain and cross-city interoperability by specifying terms from different stakeholders. Ontologies play a big role in enabling cross-city comparison. The Semantic Web has been used in the Open Government Data (OGD) approach to make it possible for cities to share information and knowledge under a common vocabulary. Pushing this further, the GCI (Global City Indicators) Ontology\cite{fox_role_2015} is an effort for the modeling of city entities that covers the concepts used by global indicators using Semantic Web technologies.

\subsection{Data Annotation}

Data can be encoded in many distinct formats including CSV, XML and NetCDF \cite{rew_netcdf:_1990}. In many cases, CSV is a format of choice because of its ease of use by both automated actors and human actors. Human actors often manually enter acquired data in a spreadsheet application (e.g., MS Excel or LibreOffice Calc). Spreadsheets are also capable of exporting content in CSV format. Basically, the CSV format can be seen as a minimalist enabling approach for data interoperability.

Regardless of the format, until our proposed CCSV format\cite{santos_contextual_2015} we are not aware that any single encoding was able to provide effective mechanisms for annotating data in a way that supports data acquisition as a contextualized data point collection. For instance, CSV lacks features for expressing the semantics associated with the data contained in it, so it is challenging to know, in an automated and interoperable way, the meaning of the data enclosed inside a CSV file. For example, it can be difficult to determine if two entries are observationally equivalent (measured under the same conditions, using the same units, in the same area, etc.). Also, different agents may generate data in different formats and standards, making CSV even more difficult to process automatically.

Although there are existing approaches for accessing CSV metadata and also for providing a metadata vocabulary for CSV data, they are typically more concerned with content restrictions, rather than the context in which the CSV data was collected. W3C's recommendations from the CSV on the Web Working Group\footnote{\url{http://www.w3.org/2013/csvw/wiki/Main_Page}} elaborate on techniques for enabling the access of CSV metadata by describing the content metadata in a separate JSON or RDF/XML file that makes use of RDF vocabulary. To bridge this gap, we proposed the Contextualized CSV (CCSV)\cite{santos_contextual_2015} as a format that deals with both content and context restrictions of the data points enclosed in it. The CCSV dataset is basically a regular CSV file with a Turtle preamble.

\subsection{Indicators}

The ISO 37120:2014\cite{iso_sustainable_2014} is a standard that defines 100 indicators across 17 themes that were evaluated to be a precise way to measure a city's performance of its services and quality of life. The themes span areas including Economy, Education, Health, and Safety. The main goal of this standard is to provide a concise set of well-defined global indicators that any city can use to measure itself. Moreover, cities that adhere to this standard are able to compare themselves, and evaluate how well they are doing in comparison to others. Making use of the ISO standard, the PolisGnosis Project\cite{fox_polisgnosis_2015} is a final goal of an ongoing effort by the University of Toronto. The project aims the following:

\begin{itemize}
\item To provide a description of all the 100 ISO indicators in terms of ontologies for the semantic web;
\item To develop an engine capable of performing analysis in order to discover root causes of differences concerning why indicators change over time for a given city and why they are different between different cities.
\end{itemize}

Until the time of this writing, the PolisGnosis Project has focused largely on the GCI Ontology engineering\footnote{\url{http://ontology.eil.utoronto.ca}} as a standard to publish the ISO indicator values, while our efforts attempt to also support a broader range of representation challenges including representation and reasoning for data visualization.

\section{City Knowledge Graph}

City indicators have some requirements that need to be followed when defining them and calculating their values. These requirements ensure that the indicator is well defined and the calculation process will generate trusted values. We have identified the following requirements:

\begin{itemize}
    \item Temporal coverage: Indicator values carry more representativeness when calculated taking into account data from a determined time frame. Such an approach enables temporal comparisons such as if a theme of interest had improving or deteriorating performance in one particular year;
    \item Entities of interest: Indicator definitions relate named entities, thus it is important to provide formal definitions for those entities;
    \item Provenance: Indicators may refer to a particular set of activities and/or data sources;
    \item Context: Indicators can also refer to data acquired under certain conditions, making context management also important;
    \item Location: Indicators values may refer to an specific area within a city or geographic region;
    \item Visualization: An easy way to visualize the calculated values is desirable.
\end{itemize}

In order for the KG to fulfill these requirements, we have made use of ontologies that can provide metadata descriptions, domain model and indicators definitions and, where the existing ontologies weren't able to cover, we have created extensions as a new ontology. The following subsections describe our choices.

\subsection{Metadata Ontologies}

City data production happens in a plethora of different sources and processes. To characterize the diversity and scale of city data produced, we reused ontologies defining the data acquisition concept, which have demonstrated their capability of encoding contextual knowledge for millions of acquired data points that would be otherwise lost during regular data acquisition activities.

\subsubsection{VSTO-I\cite{fox_ontology-supported_2009}}

"The Virtual Solar-Terrestrial Ontology - Instrument model" is an ontology\footnote{\url{http://hadatac.org/ont/vstoi}} that contains concepts that describe entities capable of collecting data (e.g., instruments, detectors and platforms) and activities related to these entities such as a deployment of an instrument on a platform. By making use of this ontology, the KG is able to keep track of all sources of data. The main reused classes are:

\begin{itemize}
\item $vstoi:Instrument$: A device, mechanism or software that is used to acquire attribute values of entities of interest.
\item $vstoi:Deployment$: A deployment is an activity of physically installing an Instrument by an agent. More than that, the deployment states that an Instrument is able to start collecting data under certain conditions (calibration, configuration etc.).
\end{itemize}

\subsubsection{HAScO}

The Human-Aware Science Ontology\footnote{\url{http://hadatac.org/ont/hasco}} is the top metadata ontology in the KG definition. HAScO describes scientific concepts related to data acquisition. With HAScO, it is possible to describe studies, projects and data collection activities like an interview of a subject or an empirical observation. HAScO is the next generation of HASNetO\cite{pinheiro_human-aware_2015} (The Human-Aware Sensor Network Ontology\footnote{\url{http://hadatac.org/ont/hasneto}}), which is a comprehensive alignment and integration of the VSTO-I sensing infrastructure and the PROV ontology. The KG makes use of the following classes, among others:

\begin{itemize}
\item $hasco:Study$: A study is a prov:Activity where steps are performed to prove or disprove an hypothesis.
\item $hasco:StudyStep$: A study step is a prov:Activity that composes a study. $hasco:DataAnalysis$ and $hasco:DataAcquisition$ are examples of it.
\end{itemize}

\subsubsection{HACitO}

\begin{figure}
\centering
\includegraphics[width=4.0in]{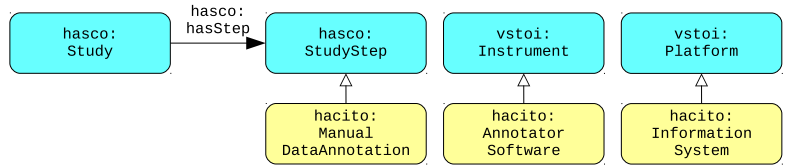}
\caption{HACitO Ontology}
\label{fig:hacito}
\end{figure}

The Human-Aware City Ontology\footnote{\url{http://hadatac.org/ont/hacito}} extends the functionalities of VSTO-I and HAScO to the Smart City context. Fig. \ref{fig:hacito} depicts the main extensions and relationships inside HACitO. HACitO makes it possible for the KG to support data production with full annotation on its origin, when the data is first generated. But, as most of the information systems in a city are legacy systems and cannot be adapted to produce fully annotated data, HACitO describes the manual data annotation data acquisition activity. The goal is to keep track of all the possible metadata involved in that data production process. For that, the ontology defines the class $hacito:ManualDataAnnotation$ as a subclass of  $hasco:StudyStep$, which is a data acquisition activity by the means of manual data annotation using an annotator software, which in turn is described by $hacito:AnnotatorSoftware$, as a subclass of $vstoi:Instrument$. The annotator is deployed to a legacy data production information system, which is an extension $hacito:InformationSystem$ of $vstoi:Platform$.

\subsection{Indicator and Domain Ontologies}

Indicators serve as metrics that provide insight into city performance. They are typically calculations over existing data. The calculated values facilitate quantitative comparisons between different cities, thus enabling city managers to make decisions informed by current data and also to support data-driven planning. The proposed KG support for indicators is based on the GCI Ontologies\cite{fox_role_2015} and the ISO 37120 Indicator Definitions Ontology\footnote{\url{http://ontology.eil.utoronto.ca/ISO37120.owl}} for the ISO 37120:2014 indicators. As discussed above, we believe that the GCI Ontology for the ISO 37120 indicators is good for publishing indicator values and comparing cities but is not aimed to support data visualization. To overcome this, the KG is able to hold user-created indicators, To address this, we have developed our QoE Indicators ontology that includes both indicators aimed at representing calculated numerical values but also indicators specifically aimed to support convenient visualization. To address this, we have developed our QoE Indicators ontology\footnote{\url{http://hadatac.org/ont/qoe}} that includes both indicators aimed at representing calculated numerical values but also indicators specifically aimed to support convenient visualization, which extends the GCI Ontology.

To make this possible, we have described the QoE indicators using the following data visualization concepts:

\begin{itemize}
\item Dimension: An entity value that usually cannot be aggregated, often used for row or columns headings;
\item Measure: An entity value that can be used to calculate something, e.g. a sum or medium, often used to support display and plotting.
\end{itemize}

\begin{figure}
\centering
\includegraphics[width=\textwidth]{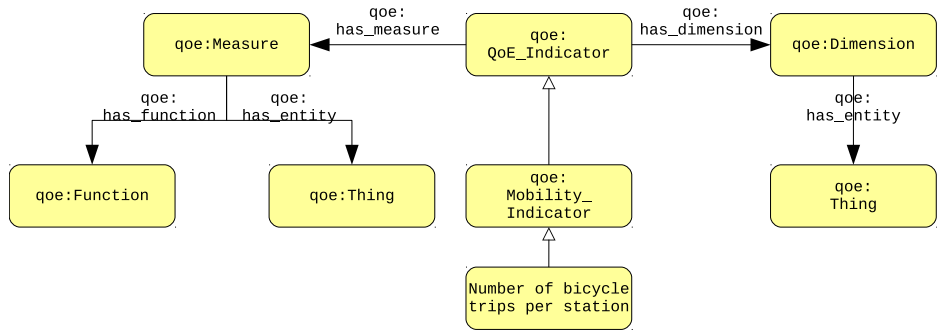}
\caption{Part of the QoE Indicators Ontology}
\label{fig:metricsANDdomain}
\end{figure}

In one example, if one has a bar chart where each bar shows the number of single commuters during a single month of the year, for a total of twelve bars, one for each month, the dimension would be the month and the measure would be sum (or count) of every person who has commuted in a given month. Fig. \ref{fig:metricsANDdomain} depicts part of the QoE Indicators Ontology. In the middle, the $qoe:QoE\_Indicator$ is defined by some $qoe:Measure$ and some $qoe:Dimension$, each of which has an associated $qoe:Thing$, i.e., the related entity. It is important also to note that the measure has a $qoe:Function$, which states what kind of calculation will be performed over that value. It is possible for an indicator to have more than one dimension and/or measure. For instance, a line chart with two measures would actually display two lines, one for each measure. An interesting case is an indicator with only measures and no dimensions. The resulting data visualization would be just a number.

\begin{figure}
\centering
\includegraphics[width=\textwidth]{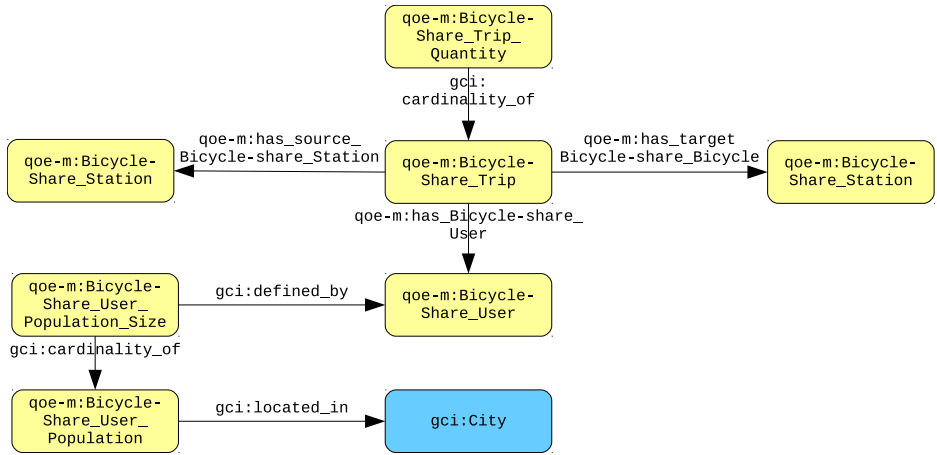}
\caption{Part of the QoE Domain Ontology}
\label{fig:domain}
\end{figure}

One of the defined indicators in the QoE Indicators Ontology is 'Number of bicycle trips per station' which defines their associated entities for dimensions and measures using the classes $qoe-m:Bicycle-Share\_Trips$ and $qoe-m:Bicycle-Share\_Station$, respectively. These domain entities are part of the QoE Domain Ontology\footnote{\url{http://hadatac.org/ont/qoe-m}} which is shown in the excerpt on Fig. \ref{fig:domain}. The QoE Ontologies (Indicators and Domain) are evolving definitions that should be tailored for each city and intended use.

\subsection{KG Serialization}

The KG serialization process is concerned with bringing the data from the KG to a physical file together with all its metadata, making it possible for third-party applications to make use of all the knowledge attached to the data. For that to take place, routines were developed to perform the following:

\begin{enumerate}
    \item A file is created for every class in the domain ontology with valid instances;
    \item For every file, write the instances as CSV registers with each column being a triple in the KG where the instance is the subject;
    \item Annotate every register and column using the CCSV format;
    \item Add to the annotations the study, deployment and data acquisitions related to the data in the file;
    \item Using the annotated data, discover suitable indicators and export them in Turtle format.
\end{enumerate}

The next section describes how this serialization was performed in the context of urban mobility.

\section{Dashboard Application}

The serialized KG is a set of files that utilize a common vocabulary, making it possible for third-party applications to interpret the CCSV format and thus understand the content and context of enclosed data. In this work, we have developed one of many possible applications: a dashboard generator. A commonly used data visualization technique is a dashboard, which is a widget that presents a number of data-based quantifications, graphs, gauges etc. Dashboards help visualize data and are closely related to instruments that humans are accustomed to using regularly. Moreover, dashboards enable human-machine interaction based on graphical visualizations, supporting a number of data analyses by the use of filters that can be applied to the data. The results of a filter application can be shown in real-time by recalculating the measures. By dynamic dashboard, we mean that the indicators displayed on the dashboard are based on the type of data presented, not predefined for a particular type. In this Section, a dashboard generating application is presented. The application is able to receive as input a serialized KG in the CCSV format together with the discovered indicators from the QoE Indicators Ontology in Turtle format to perform a metadata analysis in order to create a dashboard with as many as graphs as the presented indicators, using the QoE Indicators Ontology together with the metadata annotation to dynamically configure each visualization.

In this use case, we have worked with data acquired from the bicycle-sharing system in the city of Fortaleza, Brazil. The datasets contained data about the network formed by the usage of the system, where a user is able to grab a bicycle from a station and return it to any other station, including the station where he/she obtained it initially. We obtained two CSV files that described the network:

\begin{itemize}
    \item Bicycle-share stations: File containing only Bicycle-share stations, each station with an associated id, label and a lat/long.
    \item Trips performed: File containing all the bicycle-share system journeys. Each journey was presented with an id, an associated user that uses the bicycle, an origin and a destination bicycle-share station.
\end{itemize}

This data was collected by legacy information systems and most likely manipulated afterwards to clear up unneeded data and for better organization.

\subsection{Dataset characterization and KG manipulation}

In order to load these datasets into the KG, we first had to characterize their metadata in the following aspects:

\begin{itemize}
    \item Data source: Which ICT system or device generated this data?
    \item Data acquisition: By which data acquisition activity were they acquired? By an already able to annotate system or manual data annotation performed by an user?
    \item Study: Are the datasets part of the same study?
    \item Time frame: When were the datasets generated?
\end{itemize}

Then, we made use of tools and techniques presented in the work cited in \cite{santos_contextual_2015}, namely the CCSV format and the CCSV-Loader application. Listing \ref{serializedkg} shows part of the KG\footnote{\url{http://hadatac.org/ttl/city_kg-full.ttl}} after loading the datasets. Due to space restrictions, we present only the metadata related to the trips dataset. The data source is shown in lines 14-22 where both the annotator software and the legacy ICT system are described, while lines 1-5 states that the annotator software was deployed alongside the system at the specified date. Following, lines 6-9 describes the data acquisition activity, referring to the associated deployment. Finally, the dataset is shown in lines 10-13, where PROV-O is used to state from which activity they were generated.

\lstset{basicstyle=\small\ttfamily, columns=fullflexible, xleftmargin=5mm, framexleftmargin=5mm, numbers=left, stepnumber=1, breaklines=true, breakatwhitespace=false, numberstyle=\small, numbersep=5pt, tabsize=2, frame=lines, captionpos=b, caption={Part of the city KG}, label=serializedkg}
\begin{lstlisting}
<deployment-bss>
	a vstoi:Deployment ;
	vstoi:hasPlatform <system-bss> ;
	vstoi:hasInstrument <annotator-01> ;
	prov:startedAtTime "2016-11-08T14:42:42Z"^^xsd:dateTime .
<dataacquisition-trips>
	a hacito:ManualDataAnnotation ;
	hasco:hasContext <deployment-bss> ;
	prov:startedAtTime "2016-11-09T15:13:25Z"^^xsd:dateTime .
<dataset-trips>
	a vstoi:Dataset ;
	prov:wasGeneratedBy <dataacquisition-trips> ;
	prov:startedAtTime "2016-11-09T16:07:23Z"^^xsd:dateTime .
<annotator-01>
	a hacito:AnnotatorSoftware ;
	dc:hasVersion "X.Y"^^xsd:string .
<system-bss>
	a hacito:InformationSystem ;
	rdfs:label "Bicycle-share information system" ;
	dc:description "System for managing all the collected data from the bicycle-share system of Fortaleza."@en ;
	dc:hasVersion "X.Y"^^xsd:string ;
	dc:subject "bicycle, mobility" .
\end{lstlisting}

The following step was to serialize the KG. Listing \ref{trips_serialization} shows the CCSV preamble serialization of the $qoe-m:Bicycle-Share\_Trip$ serialization in lines 2-6. The linkage between the trip and associated stations and user are established by the station id and user id, as shown in lines 7-9. Lines 10-13 specifies the id locations for every association. The same was performed for the $qoe-m:Bicycle-Share\_Station$ entity.

\lstset{basicstyle=\small\ttfamily, columns=fullflexible, xleftmargin=5mm, framexleftmargin=5mm, numbers=left, stepnumber=1, breaklines=true, breakatwhitespace=false, numberstyle=\footnotesize, numbersep=5pt, tabsize=2, frame=lines, captionpos=b, caption={KG serialization CCSV preamble for $qoe-m:Bicycle-Share\_Trip$}, label=trips_serialization}
\begin{lstlisting}
<trips>	a vstoi:Dataset; ccsv:hasDataRecord <reg> .
<reg>
	a qoe-m:Bicycle-Share_Trip; dc:identifier <id> .
	qoe-m:has_Bicycle-Share_User <usr> ;
	qoe-m:has_source_Bicycle-share_Station <src> ;
	qoe-m:has_target_Bicycle-share_Station <trg> .
<src> a qoe-m:Bicycle-Share_Station; dc:identifier <src_id> .
<trg> a qoe-m:Bicycle-Share_Station; dc:identifier <trg_id> .
<usr> a qoe-m:Bicycle-Share_User; dc:identifier <usr_id> .
<id>      ccsv:atColumn 0 .
<src_id>  ccsv:atColumn 4 .
<trg_id>  ccsv:atColumn 7 .
<usr_id> ccsv:atColumn 1 .
\end{lstlisting}

The serialization encompasses a process for indicators discovering. The Listing \ref{prolog} shows a Prolog code we developed to verify if an indicator is suitable for the data, based on its CCSV data annotation. Lines 1-8 shows the transformation of the indicator class into Prolog rules, while lines 10-14 shows the same for the domain classes. Following, lines 16-20 shows the relations in the CCSV data annotation regarding the content of the files. In this case, the CCSV files have data records of trips and stations (line 20). The inference rules are described in lines 22-27. They make use of transitivity to verify if an indicator $Y$ is suitable for a KG $X$, i.e., if the graph contains the needed data to perform the calculation.

\lstset{basicstyle=\small\ttfamily, columns=fullflexible, xleftmargin=5mm, framexleftmargin=5mm, numbers=left, stepnumber=1, breaklines=true, breakatwhitespace=false, language=Prolog, numberstyle=\small, numbersep=5pt, tabsize=2, frame=lines, captionpos=b, caption={Prolog rules and statements for indicator discovery}, label=prolog}
\begin{lstlisting}
% indicator ontology
indicator(nr_trips_per_station).
has_value(nr_trips_per_station,nr_trips).
has_value(nr_trips_per_station,nr_station).
is_cardinality_of(nr_trips,trip).
is_cardinality_of(nr_users,user).
is_cardinality_of(nr_stations,station).
has_cardinality(X,Y) :- is_cardinality_of(Y,X).

% domain ontology
trip(trip). station(station1). station(station2).
has_user(trip,user).
has_source_station(trip,station1). has_target_station(trip,station2).
has_station(X,Y) :- has_source_station(X,Y); has_target_station(X,Y).

% dataset metadata
graph(bicicletar).
has_dataset(bicicletar,trips). has_dataset(bicicletar,stations).
dataset(trips). dataset(stations).
has_data_record(trips,trip). has_data_record(stations,station).

% inference rules
refx(X,Y) :- has_station(X,Y).
refx(X,Z) :- has_station(X,Y), refx(Y,Z).
good_ind(X,Y) :- graph(X), indicator(Y), related(X,Y).
related(X,Y) :- has_dataset(X,Z), has_data_record(Z,W), has_value(Y,V), is_cardinality_of(V,U), compatible(W,U).
compatible(X,Y) :- refx(X,Y).
\end{lstlisting}

Listing \ref{metrics} shows the discovered indicator "Trips by departure station" with its associated dimension and measure entities in Turtle format.

\lstset{basicstyle=\small\ttfamily, columns=fullflexible, xleftmargin=5mm, framexleftmargin=5mm, numbers=left, stepnumber=1, breaklines=true, breakatwhitespace=false, numberstyle=\small, numbersep=5pt, tabsize=2, frame=lines, captionpos=b, caption={Discovered indicators in Turtle}, label=metrics}
\begin{lstlisting}
<indicator01>
	a qoe:Trips_by_departure_station ;
	rdfs:label "Trips by departure station"@en ;
	qoe:dimension <dimension01> ; qoe:measure <measure01> .
<dimension01>
	a qoe:Dimension; qoe:has_entity qoe-m:Bicycle-Share_Station .
<measure01>
	a qoe:Measure; qoe:has_function qoe:Count ;
	qoe:has_entity qoe-m:Bicycle-Share_Trip .
\end{lstlisting}

\subsection{Dashboard building}

\begin{figure}
\centering
\includegraphics[width=\textwidth]{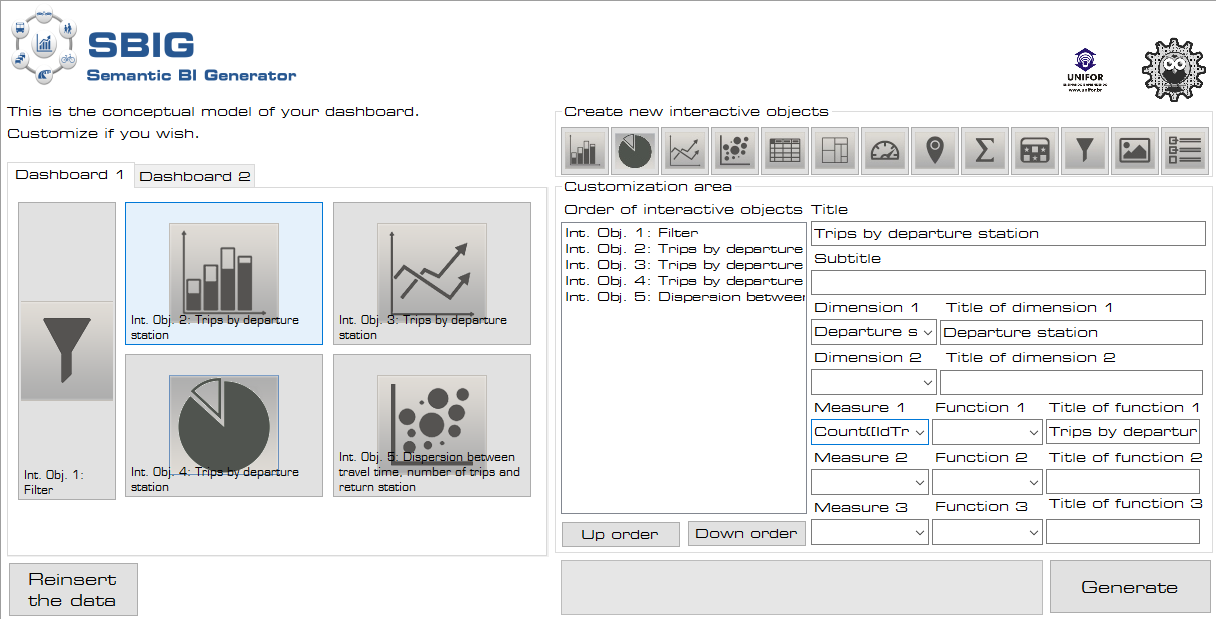}
\caption{Semantic BI Generator with the discovered indicator}
\label{fig:dash01}
\end{figure}

We have developed a dashboard generating application called the Semantic BI (Business Intelligence) Generator, which is able to interact with a number of BI solutions to automatically generate interactive dashboards based on the KG serialization. For this implementation, we focused on Qlik Sense\footnote{\url{http://www.qlik.com/us/products/qlik-sense}} which provides an API for that be used to programmatically create and setup visualizations. First, the user inputs the serialized KG and the indicators files. The tool, then, performs SPARQL queries against the indicators and KG metadata to retrieve: (i) dimension entity id column; (ii) measure entity id column; and (iii) measure calculation function. Fig. \ref{fig:dash01} shows the Semantic BI Generator after the serialized KG files and discovered metrics are loaded. On the left, a preview of the to-be-generated dashboard is presented, while on the right it is possible to modify or add new visualizations as desired. In this case, the discovered indicator is shown as a bar chart (the one currently selected), while the others have been manually added. Also, note on the right that the columns and function were filled based on the metadata information. After that, the user pushes the generate button and the tool will setup the new dashboard inside the Qlik Sense environment.

\begin{figure}
\centering
\includegraphics[width=\textwidth]{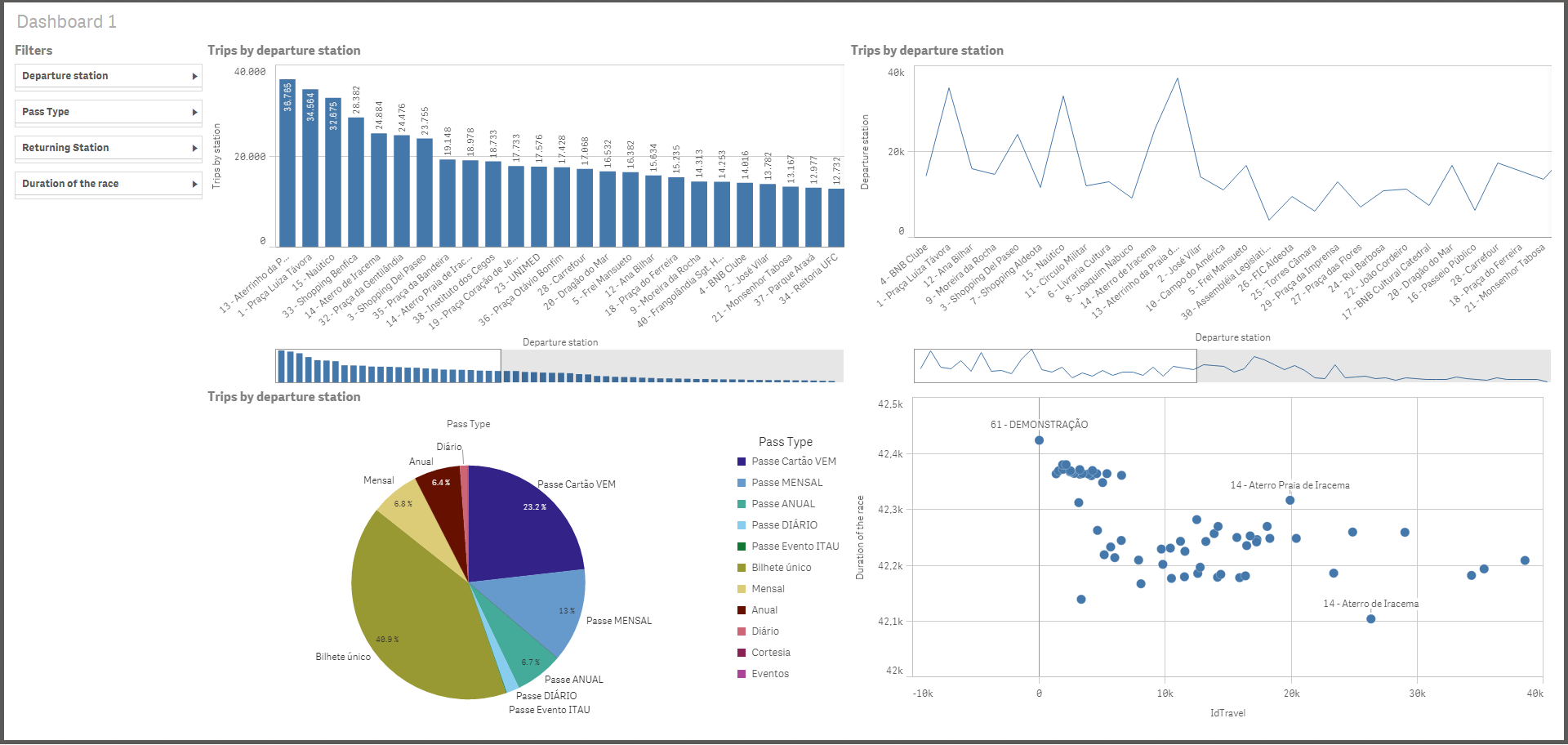}
\caption{Generated dashboard}
\label{fig:dash02}
\end{figure}

Fig. \ref{fig:dash02} shows the generated dashboard. It is possible to see the top left graph showing the dimensions as the bicycle-share stations and the measure counting the number of trips.

\section{Conclusion and Future Work}

We have presented an operational description for a city Knowledge Graph that supports automatic generation of dashboards along with an indicator ontology that supports data visualization techniques. To build our KG and to develop our indicator ontology, we have reused many existing ontologies describing identified required metadata. We also proposed an extension to the GCI Ontology focusing on data visualization concepts. A process for KG serialization with indicator discovery was performed as way to foster knowledge interoperability between the KG and third-party applications. The city KG and the QoE Ontology were used in conjunction with the Semantic Business Intelligence Generator, a dashboard generating an application capable of performing CCSV metadata analysis to automatically build rich visualizations. Potentially more importantly, the presented contributions allow users with no previous knowledge about the data (by whom and how it was generated), but who are aware of city entities and processes (that is the case for most field specialists including transportation engineers) to leverage a metadata hierarchy (provided by our ontology choices) to find the right data to be analyzed.

The research still has room to mature. For instance, we are currently working on an ontology for interactive objects to support the discovery of best suited visualization types based on an indicator definition. Also, we are continuously expanding indicators definitions to support not only data plotting but also calculation procedures (like complex network algorithms) and its associated semantics to an specific KG subset, enabling network data analytics for non-experts. In terms of KG building and metadata management, the Human-Aware Data Acquisition Framework\footnote{\url{https://tw.rpi.edu//web/project/hadatac}} (HADatAc) is being designed and developed as a framework for managing data acquired using a multitude of sources including instruments, sensors, humans, and computer models. Leveraging HAScO and VSTO-I, HADatAc is already being used in support of a number of projects, namely:

\begin{itemize}
    \item The Jefferson Project\cite{klawonn_semantic_2015}: developed in collaboration between IBM, Rensselaer Polytechnic Institute (RPI), and The FUND for Lake George;
    \item An Urban ecology project led by RPI's Center for Architecture, Science and Ecology supporting large empirical observations and a variety of experiments.
    \item The Smart City Center at the Universidade of Fortaleza where scientific observations are conducted to understand the use of city resources in support of mass transportation.
    \item The CHEAR\footnote{\url{https://tw.rpi.edu//web/project/CHEAR}} Project where ontologies are being developed to support research on exposure science and child health and also tools and infrastructure for building and maintaining a knowledge graph of related content.
\end{itemize}

\bibliographystyle{splncs03} 
\bibliography{ESWC-17}

\end{document}